\documentclass[10pt,twocolumn,letterpaper]{article}

\usepackage[pagenumbers]{cvpr} 

\definecolor{cvprblue}{rgb}{0.21,0.49,0.74}
\usepackage[pagebackref,breaklinks,colorlinks,allcolors=cvprblue]{hyperref}

\def\paperID{11702} 
\def\confName{CVPR}
\def\confYear{2026}

\title{Beyond Perfect Scores: Proof-by-Contradiction for Trustworthy Machine Learning}

\author{Dushan N. Wadduwage\textsuperscript{*}\\
Old Dominion University\\
Norfolk, VA, USA\\
{\tt\small dwadduwa@odu.edu}\\
{\small \textsuperscript{*}Corresponding author}
\and
Dineth Jayakody\\
Old Dominion University\\
Norfolk, VA, USA\\
{\tt\small dinethjayakody@cs.odu.edu}
\and
Leonidas Zimianitis\\
Old Dominion University\\
Norfolk, VA, USA\\
{\tt\small lzimi001@odu.edu}
}

\begin{document}

\newcommand{\dnw}[1]{\textcolor{red}{[dnw:#1]}}

\newcommand{\dinj}[1]{\textcolor{Green}{[dinj:#1]}}
\newcommand{\leo}[1]{\textcolor{Yellow}{[leo:#1]}}

\maketitle
\begin{abstract}

Machine learning (ML) models show strong promise for new biomedical prediction tasks, but concerns about trustworthiness have hindered their clinical adoption. In particular, it is often unclear whether a model relies on true clinical cues or on spurious hierarchical correlations in the data. This paper introduces a simple yet broadly applicable trustworthiness test grounded in stochastic proof-by-contradiction. Instead of just showing high test performance, our approach trains and tests on spurious labels carefully permuted based on a potential outcomes framework. A truly trustworthy model should fail under such label permutation; comparable accuracy across real and permuted labels indicates overfitting, shortcut learning, or data leakage. Our approach quantifies this behavior through interpretable Fisher-style $p$-values, which are well understood by domain experts across medical and life sciences. We evaluate our approach on multiple new bacterial diagnostics to separate tasks and models learning genuine causal relationships from those driven by dataset artifacts or statistical coincidences. Our work establishes a foundation to build rigor and trust between ML and life-science research communities, moving ML models one step closer to clinical adoption.

\end{abstract}    
\section{Introduction}
\label{sec:intro}

Machine learning approaches have revolutionized multiple facets of biomedical and clinical prediction models. However, 
the generalizability of some ML-based prediction models remains questionable. \citet{chekroud2024illusory}
recently systematically demonstrated that a model trained on data from one clinical study did not generalize to data from other similar studies, in multiple cases. Currently, such failure modes cannot be detected until multiple datasets from multiple studies become available. A high-performing prediction model may mislead researchers to invest millions of dollars of resources in a seemingly promising research direction that may later fail. Thus, a scientifically rigorous technique to evaluate generalizability early on is needed.

\begin{figure}
    \centering
    \includegraphics[width=1.0\linewidth]{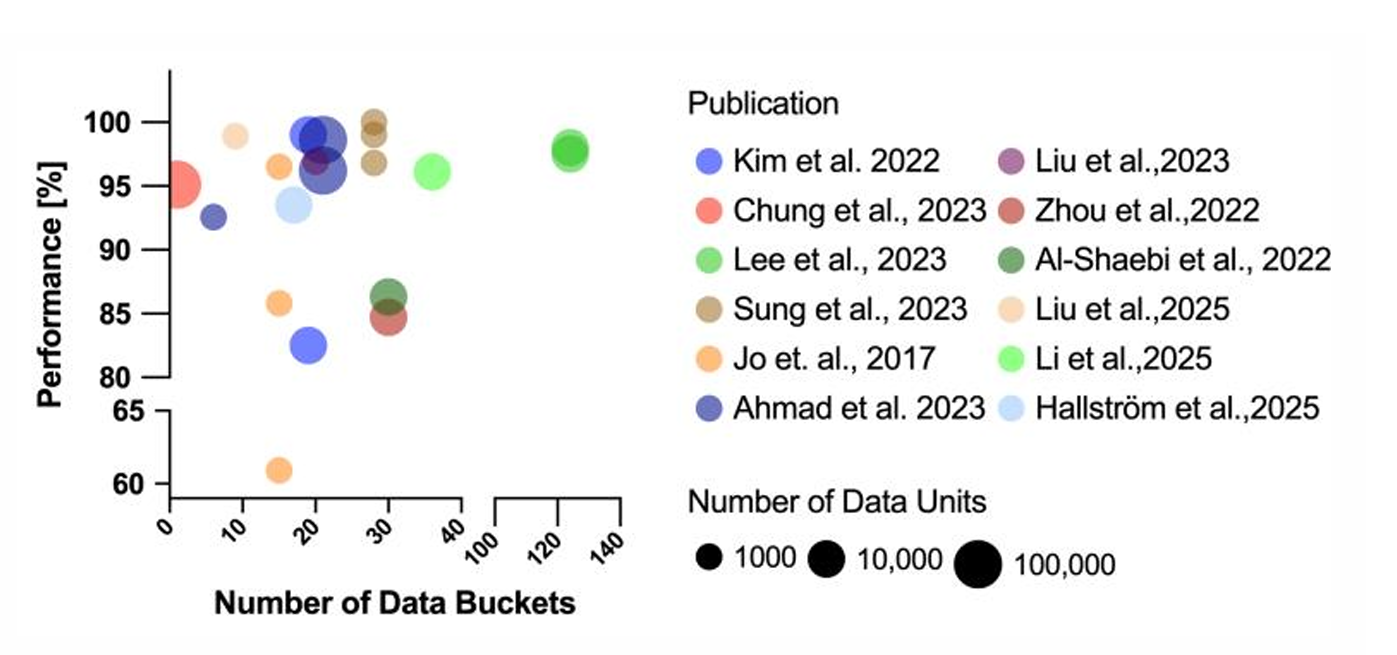}
    \caption{Representative papers with highly accurate deep
learning models trained with small numbers of data buckets~\cite{chung2023label,kim2022rapid,lee2023machine,sung2023three,jo2017holographic,ahmad2023highly,hallstrom2025rapid,li2021deep,nissim2021real,al2022highly, liu2023classification,liu2025deep,zhou2022ramannet}  }
    \label{fig:bucketgraph}
\end{figure}

Models lack generalizability when they overfit to the training data. Overfitting can be readily detected through well-established “train-validate-test” guidelines with enough independent and identically distributed (IID) data. But when the IID assumption is violated, generalizability conclusions from the standard train-validate-test method are questionable. In general, the IID assumption is related to the data generation (or sampling) process; for scientific and medical data, a domain expert should decide if the data is IID enough before an ML model is utilized, so that the violation can be addressed during model training. However, this critical step is notoriously difficult to practice, and the IID condition is often simply assumed.

We identified a common hierarchy in biomedical data that structurally violates the IID assumption. Large biomedical and clinical datasets often originate from only a few subjects. For example, in a mouse study, millions of cells (data units) can be measured, but they often originate from tens of mice (data buckets). In a pathogen study, millions of bacteria can be measured, but they often originate from tens of clinical isolates or strains. Similarly, in a clinical trial, thousands of data units may be collected, but they often originate from tens of patients. In all such situations, data units from the same data bucket may very well be correlated, violating the IID assumption. Many biomedical datasets seem to have been curated and used without accounting for this IID violation. For instance, researchers in quantitative phase microscopy (QPM) and Raman spectroscopy, which are two emerging label-free techniques with a broad range of biomedical and clinical applications~\cite{park2023artificial}, frequently use ML prediction models. But in many cases, their training datasets are collected from tens of independent specimens. Fig.~\ref{fig:bucketgraph} shows a set of published Phase microscopy~\cite{chung2023label,kim2022rapid,lee2023machine,sung2023three,jo2017holographic,ahmad2023highly,hallstrom2025rapid,li2021deep,nissim2021real} and Raman spectroscopy~\cite{al2022highly, liu2023classification,liu2025deep,zhou2022ramannet} studies that used thousands to hundreds of thousands of data units to train ML models with impressive near-100\% accuracy. But these data units originated from a much fewer data buckets (see the x-axis in Fig.~\ref{fig:bucketgraph}). Our results (in Section~\ref{sec:Evaluation}) suggest that some of these models likely suffer from non-IID sampling and may not be trustworthy. 
 
This paper proposes a simple yet broadly applicable test to evaluate model generalizability in the presence of possible hierarchical non-IID data. Our method is grounded on the Rubin causal model (RCM)~\cite{imbens2008rubin} (also known as the potential outcomes framework). Instead of just showing superior test performance, our approach trains and tests on spurious labels carefully permuted by RCM, based on the data hierarchy. If the models trained on spurious labels perform as well as the original, the original model is unreliable despite high accuracy. As a metric of trustworthiness, our approach calculates Fisher-style p-values ~\cite{bind2020possible}, which are well understood by domain experts across medical and life sciences. Thus, our contributions will lay the very foundation to build rigor and trust between ML and life-science research communities.
  
\noindent{Our main contributions are as follows:}
\begin{itemize}
    \item We systematically identify and formalize a pervasive non-IID sampling structure in biomedical datasets, termed the many-units–few-buckets phenomenon , which challenges the validity of conventional train–validate–test evaluation pipelines.
    \item We propose a framework grounded in the Rubin causal model that quantifies model trustworthiness through a permutation-based stochastic proof-by-contradiction, offering a principled statistical test for trustworthiness.
    \item We introduce a universal, interpretable, and dataset-agnostic benchmark for evaluating ML reliability across diverse biomedical imaging domains, providing a standardized measure of trustworthiness beyond conventional accuracy metrics.
\end{itemize}

\section{Related Work}
\label{sec:Related Work}

Many biomedical ML applications report excellent internal performance, yet audits show these scores often overstate \emph{trustworthy} generalization. One concern is hidden violations of independence in evaluation. \citet{tampu2022inflation} show that splitting volumetric OCT scans at the per-image level induces train--test leakage because adjacent B-scans from the same volume are near-duplicates, inflating accuracy by up to 30\% versus subject/volume-level splits. \citet{kapoor2023leakage} review ML applications and document 294 papers with data leakage, including duplicate or near-duplicate samples across splits and temporal or sampling biases that misalign evaluation and target populations. These works suggest that standard train--validate--test pipelines under unexamined IID assumptions can mislead about reliability.

A second line of work traces these failures to \emph{shortcut learning}, where models exploit cues that correlate with labels only in a narrow regime but are semantically misaligned. \citet{geirhos2020shortcut} argue that deep networks follow a “path of least resistance,” e.g., textures or context rather than object shape, and thus generalize poorly under distribution shifts. In chest radiography, \citet{zech2018variable} show pneumonia detectors trained on one hospital’s X-rays achieve strong internal AUC but degrade on external hospitals because they learn hospital-specific signatures instead of pathology. Similarly, \citet{degrave2021ai} find radiographic COVID-19 detectors with near-perfect internal AUC rely on laterality markers, projection type, or institution-specific processing instead of genuine pathology. Explanatory analyses such as \citet{lapuschkin2019unmasking} uncover similar “Clever Hans” behavior, where high test accuracy is driven by dataset artifacts rather than the intended signal. Together, these studies show that high i.i.d.\ accuracy alone is insufficient to guarantee clinically meaningful behavior, especially when data exhibit strong, but often unmodeled, hierarchical structure across buckets such as sites, patients, or specimens.

Several works define richer notions of \emph{trustworthiness} for biomedical ML. In medical imaging, the consensus statement by \citet{aldieri2025consensus} conceptualizes the \emph{credibility} of ML predictors as knowing prediction error across a specified information space and advocates a seven-step assessment covering context of use, uncertainty quantification, error decomposition, and robustness to biases. At the prediction level, \citet{nicora2022evaluating} review methods for assessing the reliability of predictions and introduce a classifier-agnostic framework combining a \emph{density principle} (flagging inputs far from the training distribution) with a \emph{local fit principle} (checking whether the model was accurate on nearby training points), enabling selective rejection of unreliable cases. Socio-technical work by \citet{zicari2021assessing} introduces the Z-Inspection\textregistered{} process to assess a deployed OHCA detection system, surfacing issues around alert fatigue, fairness, accountability, and continuous monitoring. While these frameworks move beyond raw accuracy, they require problem-specific analyses or system-level auditing and do not offer a \emph{universal statistical test} for arbitrary biomedical classifiers to check whether reported performance is compatible with genuine learning under hierarchical data generation.

Causal inference offers a principled lens on robustness and interpretability in ML, including under non-IID sampling. Surveys such as \citet{guo2020survey} and \citet{jiao2024causal} review how structural causal models and potential-outcomes frameworks guide effect estimation and causal discovery across data regimes with interference, networks, and temporal dependence. In healthcare, \citet{sanchez2022causal} argue that embedding causal structure into clinical ML and precision medicine can mitigate shortcut learning and improve out-of-distribution performance. Closer to our setting, \citet{zhang2023causal} analyze causal estimation under non-IID sampling induced by interference between units with partially known interaction graphs; they quantify bias from naively assuming IID, propose subset-selection-based debiasing, and characterize identifiability of a “true” causal effect. 

Non-IID structure also arises in federated learning (FL), where sites or devices act as natural buckets. \citet{shae2022thoughts} argue that the “more data is better” intuition can fail in healthcare FL because cross-hospital datasets are typically non-IID, and propose metrics and clustering strategies to quantify and accommodate inter-site skews. \citet{jimenez2025thorough} empirically quantify non-IIDness via Hellinger distance across label, feature, quantity, and spatiotemporal skews, showing how these skews degrade FL accuracy and recommending routinely reporting non-IID metrics. These works make non-independence explicit at the level of networks or sites and provide tools for learning with complex sampling, but they largely assume the hierarchy is known and target model design and aggregation rather than a general-purpose hypothesis test for the trustworthiness of a given trained classifier.

Our work also connects to classifier-based hypothesis testing and permutation-based inference. The Classifier Two-Sample Test (C2ST) of \citet{friedman2004multivariate} reframes distribution comparison as binary classification and uses hold-out accuracy as a test statistic to detect differences between two i.i.d.\ samples. Permutation tests for classifier performance \citep{ojala2010permutation} estimate empirical $p$-values by comparing observed error to a null distribution obtained by permuting labels (or features) under IID sampling. Extending such ideas, \citet{mi2021permutation} propose PermFIT, a model-agnostic feature-importance test that uses permutation-based error increases and cross-fitting to obtain per-feature $p$-values without refitting the base learner. In positive–unlabeled learning, \citet{xu2024leveraging} couple PU-bagging with a permutation test over P/U labels to decide whether predictions meaningfully generalize to unlabeled data. These approaches bring nonparametric hypothesis testing closer to modern ML but generally assume IID samples and treat permutations at the individual-example or feature level rather than respecting a hierarchical data-generation process.

In contrast to prior work, we make the \emph{hierarchy of biomedical data generation} explicit and use it as the backbone of our evaluation procedure. Many biomedical studies collect thousands to millions of data units (cells, spectra, image patches) from only tens to hundreds of buckets (patients, specimens, sites, time points), but this many-units--few-buckets structure is rarely formalized in ML evaluations. Building on the potential-outcomes framework and classifier-based tests such as C2ST, we construct a Rubin-style causal null in which bucket-level labels are stochastically permuted and test whether a model’s observed accuracy can be explained by this null. If a model continues to perform well under bucket-level label permutations, its apparent success can be attributed to hierarchical or other non-IID artifacts rather than genuine learning of label-relevant features. Quantifying evidence against this null through Fisher-style permutation $p$-values yields a universal, model- and dataset-agnostic notion of \emph{causal significance} that complements existing trustworthiness frameworks, causal ML methods, and permutation-based tests and directly targets the many-units--few-buckets phenomenon pervasive in biomedical imaging.

\section{Methodology}
\label{sec:Methodology}


\begin{figure}[t]
    \centering
    \includegraphics[width=\columnwidth]{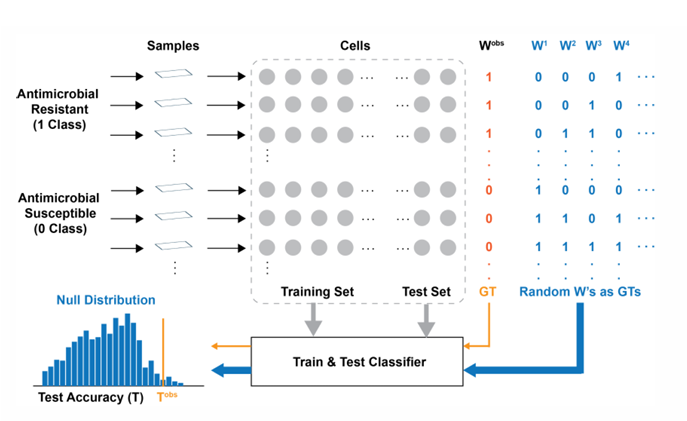}
    \caption{
    Overview of the proposed causal null-model framework for evaluating model trustworthiness through permutation-based stochastic proof-by-contradiction.
    Observed bucket-level labels $\mathbf{W}^{\text{obs}}$ serve as ground truth (GT).
    Random permutations $\{\mathbf{W}^{(m)}\}_{m=1}^M$ act as alternative ground truths to form the null distribution of test accuracies.
    The Fisher-style $p$-value quantifies whether the observed model performance $T_{\text{obs}}$ is significantly higher than those obtained under random label assignments.
    }
    \label{fig:framework}
\end{figure}


Before we introduce the formalism, we illustrate our main idea through an example. Consider a new ML-based diagnostic for antimicrobial resistance (AMR). As shown in Fig.~\ref{fig:framework}, we may image bacteria from a group of resistant samples and compare their images to those from an antimicrobial-susceptible group of samples. Note that each sample contains many bacteria with the same label (either resistant or susceptible). Thus, even with a few samples (data buckets), one can generate many individual data points (data units) to train an ML classifier. Typically, if the trained model accurately predicts resistant vs. susceptible on the test data, we would trust the model. Yet this conclusion is only valid if the data within the buckets aren’t correlated due to reasons other than AMR. 

In our proposed framework, rather than simply trusting the test accuracy, we treat the classification process as a hypothesis test, with our test accuracy as the test statistic. Our preferred hypothesis is that the bacterial images contain an underlying signature of AMR that can be detected by a machine learning classifier. The corresponding null hypothesis asserts the absence of such a signature, meaning the class labels contain no information. Under the null hypothesis, if we randomly permute the bucket-level class labels (denoted as permutations $\{W_1, W_2, W_3, \dots\}$ in Fig.~\ref{fig:framework}) and train a classifier on each permuted dataset, those classifiers should perform similarly to the original classifier trained on the true labels. Repeating this experiment across many random permutations allows us to generate a histogram of accuracy values that approximate the null distribution of the test statistic. The Fisher-style $p$-value, computed as the percentile rank of the observed classifier’s performance, $T_{\text{obs}}$, quantifies the evidence against the null hypothesis. Higher p-values indicate that the null hypothesis of no signature cannot be rejected irrespective of the performance of the original classifier. 

Next, we formalize our approach and review essential statistical tools for causal inference and permutation testing.

\subsection{Classification with Bucket-Structured Data}
Let $\mathcal{D} = \{(\mathbf{x}_i, y_i)\}_{i=1}^N$ be a labeled dataset, where $\mathbf{x}_i \in \mathcal{X}$ (e.g., an image or spectrum) and $y_i \in \mathcal{Y} = \{1, \dots, K\}$ is a class label ($K \geq 2$). 
The data are grouped into $B$ independent biological \emph{buckets} (e.g., patients, mice, or isolates):
\begin{equation}
\mathcal{D} = \bigcup_{b=1}^{B} \mathcal{D}_b, \quad 
\mathcal{D}_b = \{(\mathbf{x}_i, y_i) : i \in \mathcal{I}_b\},
\end{equation}
where $\mathcal{I}_b$ denotes the \emph{index set} of samples belonging to bucket $b$, such that each $\mathcal{I}_b \subseteq [N] = \{1, 2, \dots, N\}$. 
With $\mathcal{I}_b \cap \mathcal{I}_{b'} = \emptyset$ and $\bigcup_b \mathcal{I}_b = [N]$, all units within a bucket share the same class label: $y_i = W_{b(i)}$, where $b(i)$ maps unit $i$ to its bucket and $W_b \in \mathcal{Y}$ is the bucket-level label. 
This hierarchical structure induces intra-bucket correlation, violating the IID assumption and leading to inflated accuracy in standard train--validate--test splits.

A classifier $f_\theta: \mathcal{X} \to \mathcal{Y}$ is trained to minimize the empirical
 risk on a training partition. 
Performance on a test set $\mathcal{S}$ is evaluated using a chosen test statistic $T(\mathcal{S})$, such as accuracy:
\begin{equation}
T(\mathcal{S}) = \text{Acc}(f_\theta, \mathcal{S}) 
= \frac{1}{|\mathcal{S}|} 
\sum_{(\mathbf{x}, y) \in \mathcal{S}} 
\mathbb{I}\!\left[ f_\theta(\mathbf{x}) = y \right].
\end{equation}

Although accuracy is sensitive to class imbalance, this does not invalidate the permutation test because the null distribution is generated under the same imbalanced data structure. The test remains valid for detecting whether the model exploits systematic bucket-level signal beyond chance, even if raw accuracy is inflated.


\subsection{Causal Null Hypothesis via the Rubin Model}
Following the Rubin causal model, we treat the bucket label $W_b$ as a treatment assignment applied to all units in bucket $b$. 
The causal null hypothesis ($H_0$) states that, conditional on bucket identity, the features contain no class-specific predictive signal:
\begin{equation}
H_0: \mathbf{x}_i \perp\!\!\!\perp W_{b(i)} \mid b(i).
\end{equation}
where $\perp\!\!\!\perp$ denotes statistical independence and $b(i)$ maps each sample to its corresponding bucket.
Under $H_0$, reassigning bucket labels should not increase accuracy beyond chance.

\subsection{Permutation Test}
To test $H_0$, we perform bucket-level label permutation. Let $\mathbf{W}^{\text{obs}} = (W_1, \dots, W_B)$ be the observed bucket labels. A random permutation $\pi: [B] \to [B]$ generates:
\begin{equation}
\mathbf{W}^{(\pi)} = (W_{\pi(1)}, \dots, W_{\pi(B)}), \quad y_i^{(\pi)} = W_{\pi(b(i))}.
\end{equation}
The permuted dataset is $\mathcal{D}^{(\pi)} = \{(\mathbf{x}_i, y_i^{(\pi)})\}_{i=1}^N$. 
Under $H_0$, the distribution of accuracy $T(\mathcal{D}^{(\pi)})$ is exchangeable across all permutations $\pi$.

Under the alternative hypothesis ($H_1$), the features $\mathbf{x}_i$ carry genuine class-specific signal, such that the observed accuracy $T_{\text{obs}} = T(\mathcal{D})$ is expected to be significantly higher than those obtained under random label permutations. 
Rejecting $H_0$ in favor of $H_1$ thus indicates that the classifier has captured a true, generalizable relationship between the features and the bucket-level labels.

\subsection{Fisher-Style $p$-Value via Permutation Testing}

To quantify statistical significance, we use Fisher's exact randomization principle.
Under the null hypothesis $H_0$, permuting the bucket-level labels does not affect the test statistic.
Let $\Pi_B$ denote the set of all possible label permutations across $B$ buckets.
For each permutation $\pi \in \Pi_B$, we train a classifier on the permuted dataset $\mathcal{D}^{(\pi)}$ and compute its test statistic:
\begin{equation}
T_\pi = T(\mathcal{D}^{(\pi)}).
\end{equation}
The observed statistic on the true labels is $T_{\text{obs}} = T(\mathcal{D})$.
The Fisher-style $p$-value is defined as:
\begin{equation}
p = \frac{\sum_{\pi \in \Pi_B} \mathbb{I}\!\left[T_\pi \geq T_{\text{obs}}\right]}{|\Pi_B|}.
\end{equation}
where $\mathbb{I}[\cdot]$ is the indicator function that equals $1$ if the condition inside the brackets is true and $0$ otherwise.
This measures the probability, under random bucket-level label assignments, of obtaining a test statistic at least as large as the observed one.
A small $p$-value (e.g., $p < 0.05$) rejects $H_0$, indicating that the classifier’s performance cannot be explained by chance and thus reflects a generalizable signal.
In practice, since the number of unique label permutations $M = |\Pi_B|$ is finite, the $p$-value is discrete, and its minimum attainable value is $p_{\min} = 1/M$.
This lower bound determines the resolution of statistical significance, that results with $p$ approaching $p_{\min}$ indicate the strongest possible evidence against $H_0$ given the available number of permutations.

\section{Evaluation}
\label{sec:Evaluation}

We evaluate our proposed framework on both controlled benchmarks and published biomedical datasets. The toy benchmarks validate the framework’s expected behavior based on a simulated visual cue (and the lack thereof), while the biomedical experiments demonstrate its utility for evaluating trustworthiness in real-world scenarios.

\subsection{Controlled Benchmarks}
\label{sec:controlled}

We design controlled benchmarks in which the strength of the underlying causal signal can be explicitly manipulated. We consider two datasets, Rotated MNIST and Colored FashionMNIST, each constructed to introduce visual cues under well-defined conditions. Across all experiments, we employ two representative architectures: a lightweight convolutional neural network (\emph{LightCNN}) and a compact Vision Transformer (\emph{LightViT}). The \emph{LightCNN} serves as a strong inductive baseline that captures local spatial structure, while the \emph{LightViT} provides an attention-based counterpart capable of modeling global dependencies. Both models are used consistently across all experiments to establish controlled and reproducible benchmarks for evaluating the proposed framework (model details described in Supplementary Section).

\subsubsection{Rotated MNIST}

We evaluate whether a rotation cue enables binary discrimination on the MNIST dataset.
Digits \(\{0,1,2,3,4\}\) are defined as \emph{treated} (T) and digits \(\{5,6,7,8,9\}\) as \emph{untreated} (F).
For each experiment, a per-class rotation angle is sampled from truncated Gaussian distributions
(\(T \sim \mathcal{N}(\theta, \sigma=2^\circ)\), \(F \sim \mathcal{N}(0^\circ, \sigma=2^\circ)\)),
clipped to the interval \([-90^\circ, 90^\circ]\).
Each image of a given class is rotated by its corresponding class angle using bilinear interpolation (no expansion, fill value = 0). We sweep treatment angles $\theta \in \{0^\circ\!:\!5^\circ\!:\!30^\circ\} \cup \{45^\circ, 60^\circ, 90^\circ\}$
, using a unique random seed for angle generation and a fixed seed for deterministic training. We perform a permutation test for all unique class-to-label assignments.
For each permutation, the models were trained for 5 epochs to produce a null accuracy distribution,
from which the \(p\)-value is computed.

\begin{figure}[!t]
  \centering

  \begin{subfigure}[t]{0.22\textwidth}
    \centering
    \includegraphics[width=\linewidth]{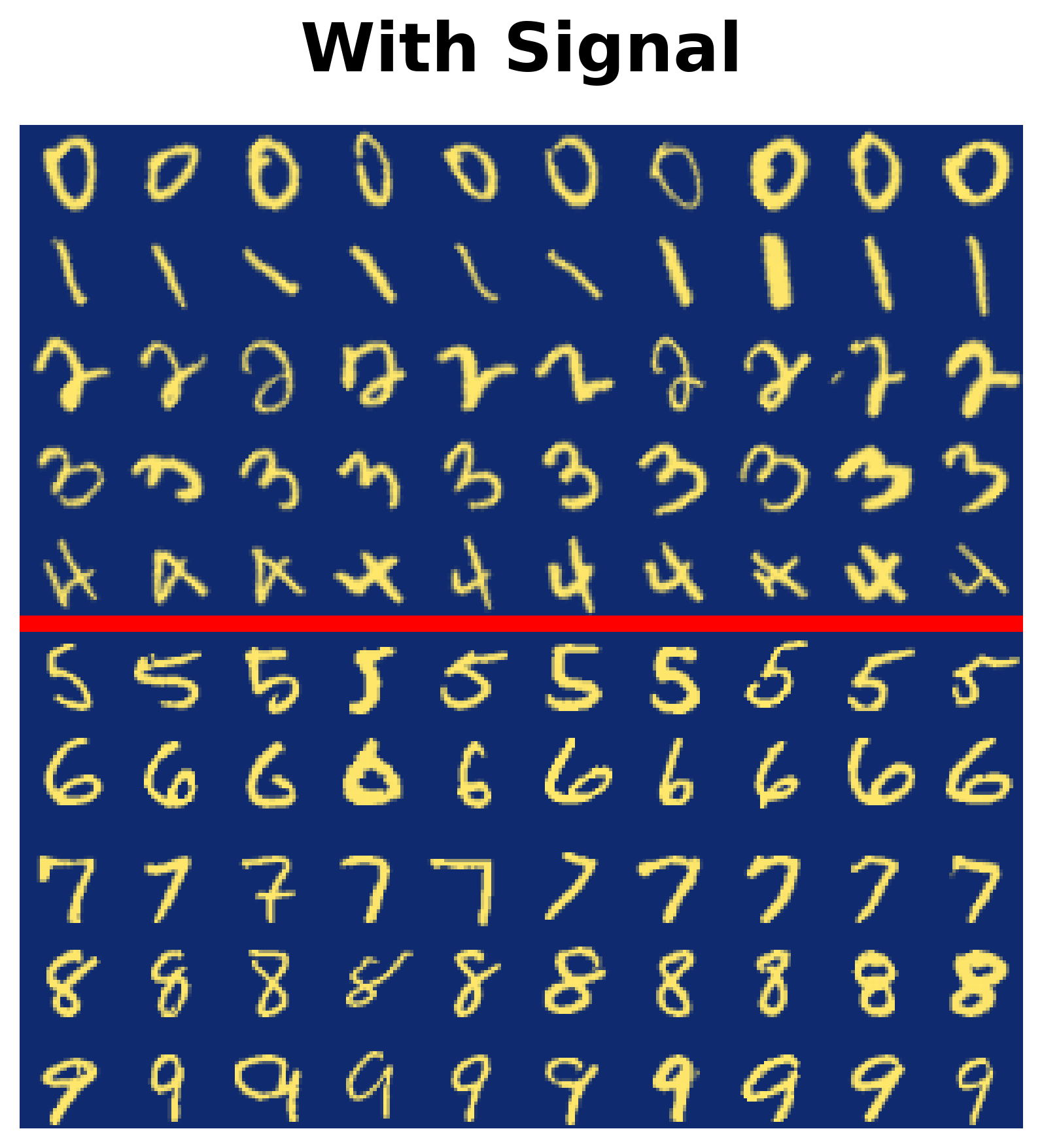}
    \caption{}\label{fig:a}
  \end{subfigure}\hfill
  \begin{subfigure}[t]{0.22\textwidth}
    \centering
    \includegraphics[width=\linewidth]{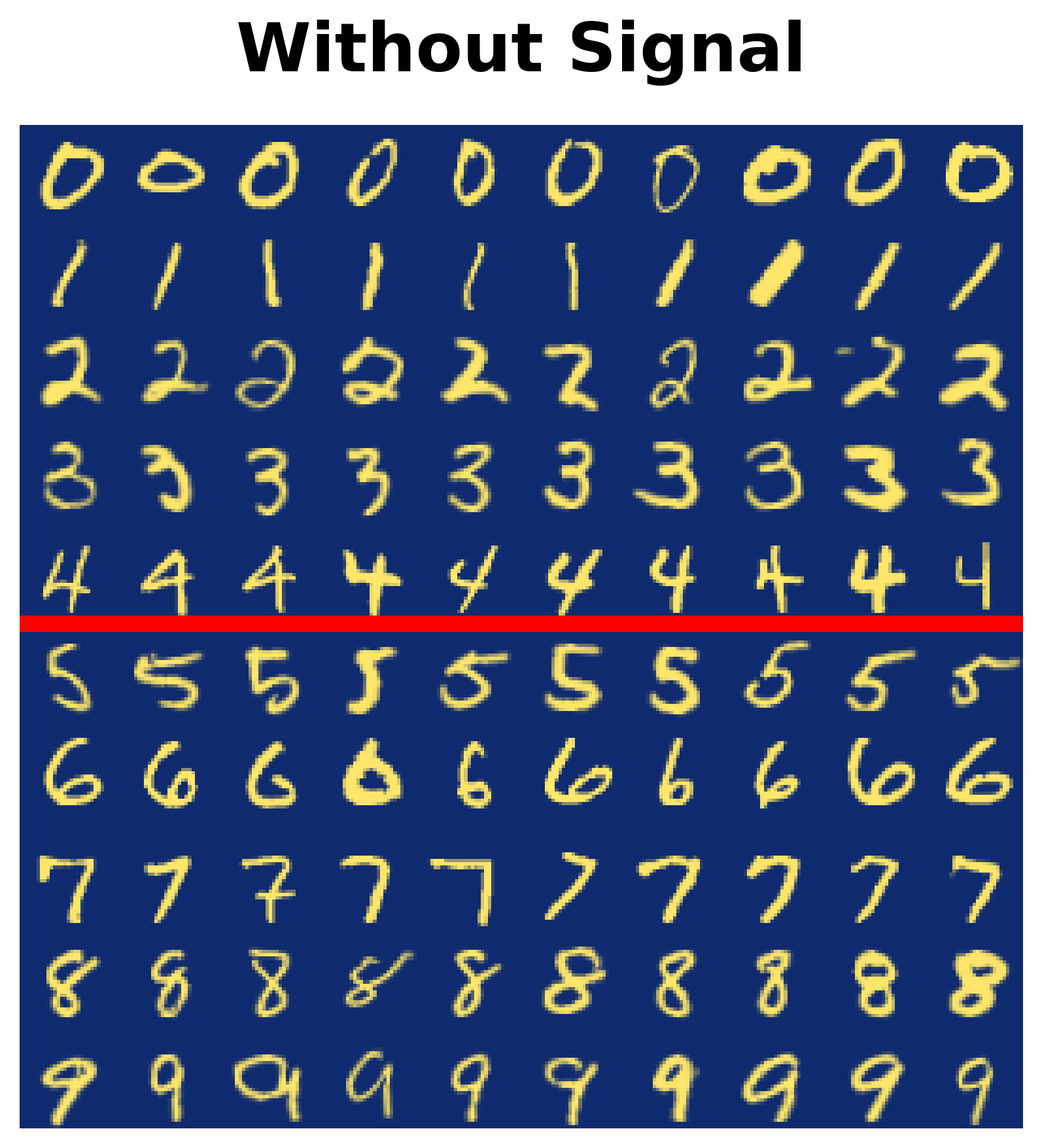}
    \caption{}\label{fig:b}
  \end{subfigure}

  \vspace{0.6em}

  \begin{subfigure}[t]{0.23\textwidth}
    \centering
    \includegraphics[width=\linewidth]{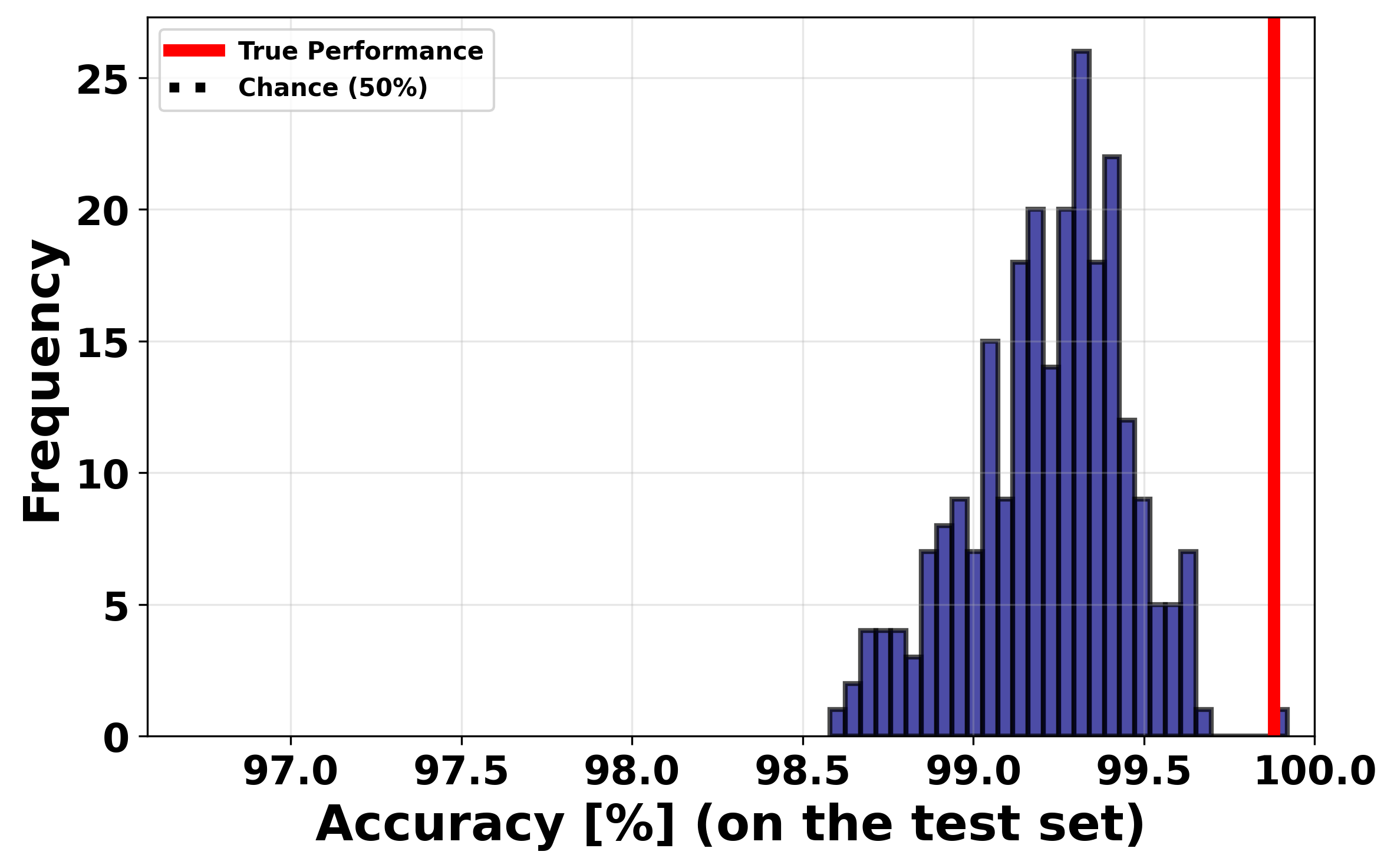}
    \caption{}\label{fig:c}
  \end{subfigure}\hfill
  \begin{subfigure}[t]{0.23\textwidth}
    \centering
    \includegraphics[width=\linewidth]{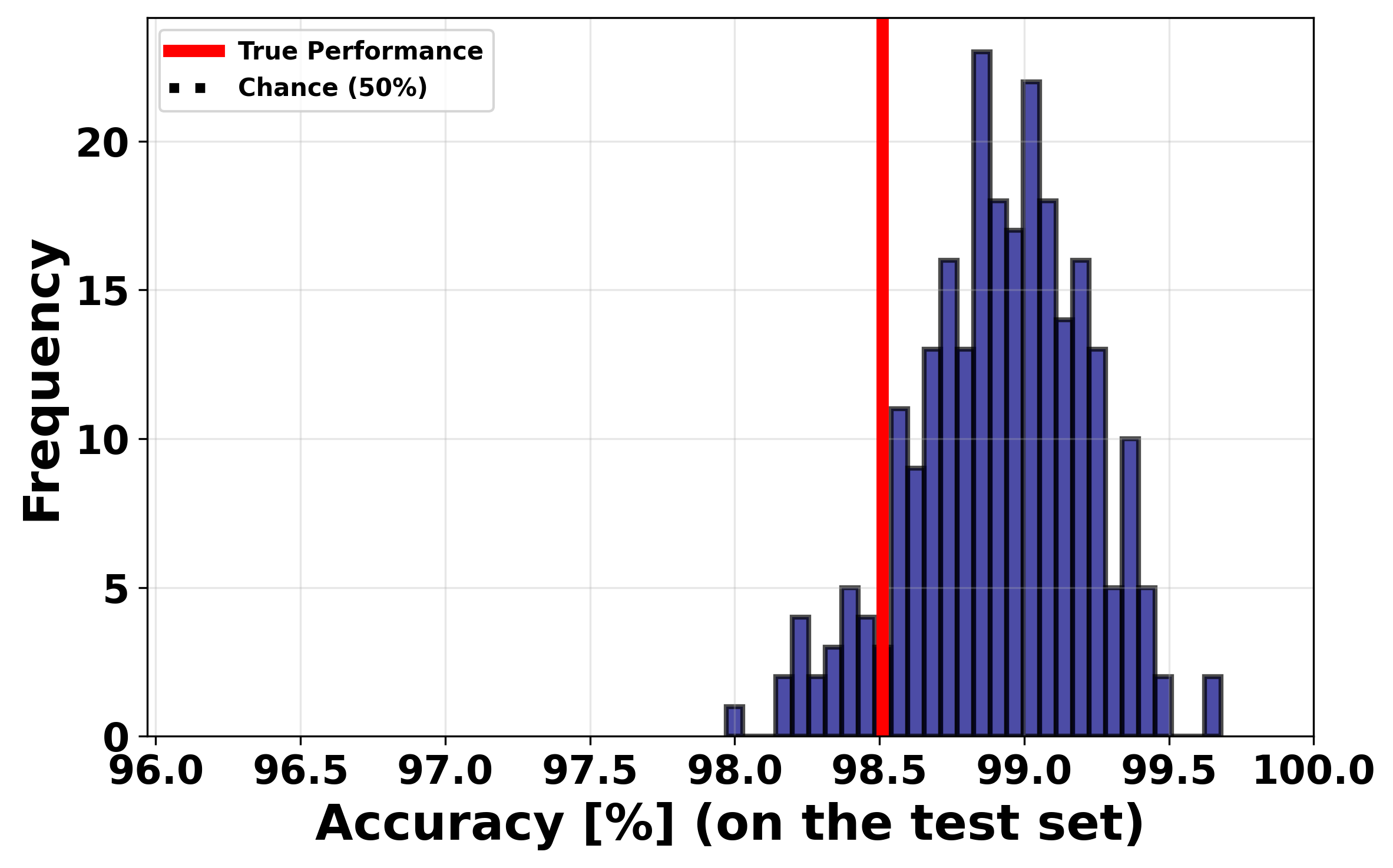}
    \caption{}\label{fig:d}
  \end{subfigure}

   \caption{Simulated MNIST example with and without a causal signal on LightCNN. 
  (a) Digits rotated by \(45^\circ\) introduce a treatment cue, yielding significant separation between observed and null accuracies. 
  (b) Without rotation, no true signal exists despite high apparent accuracy.}
  \label{fig:mnist_comparison}
\end{figure}

\begin{figure*}[!t]
    \centering
    \includegraphics[width=1.0\textwidth]{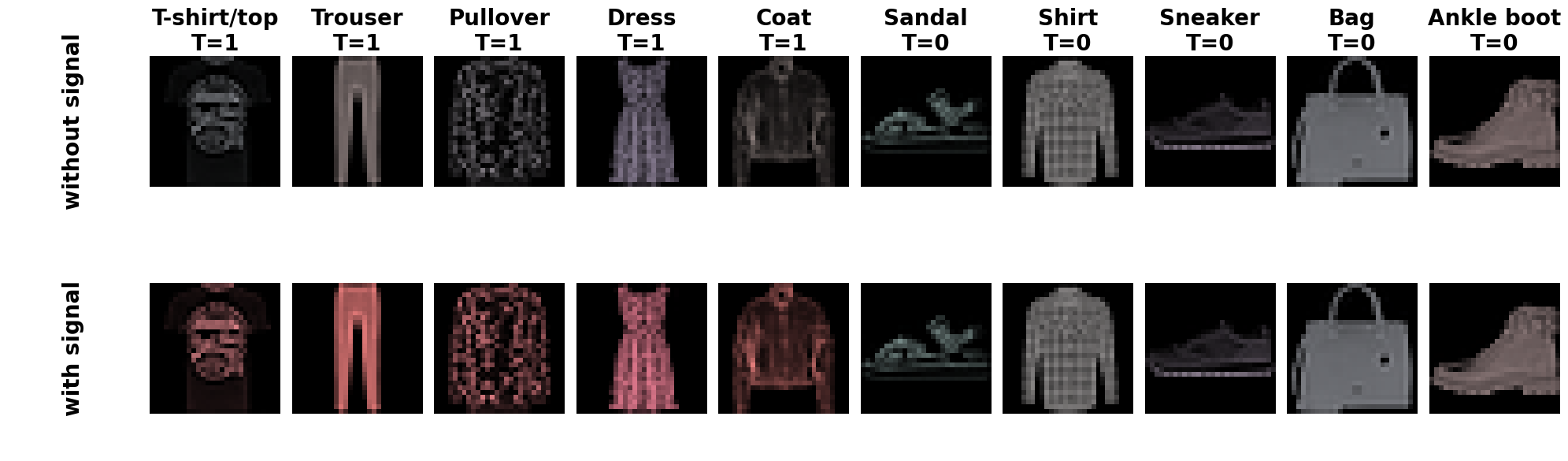}
    \caption{Sample images Top: RGB versions of the images without any class-dependent color manipulation; 
Bottom: RGB images with class-dependent red scaling and shared green/blue baselines, altering global color without changing spatial structure.
}
    \label{fig:fashionmnist}
\end{figure*}

As shown in Fig.~\ref{fig:mnist_comparison}, 
when no signal was introduced (\(0^{\circ}\) rotation), the \emph{LightCNN} still achieved a high apparent accuracy (\(\sim99\%\)) but with a high \(p\)-value (\(p=0.9685\)), demonstrating that accuracy alone can be misleading as a measure to trust the model detecting an expected cue. 
In contrast, when \(45^{\circ}\) rotation was applied, the p-value dropped close to the theoretical minimum (\(p=0.0080\)) along with a slight increase in the model accuracy. Results for additional rotation angles are summarized in Table~\ref{tab:causal_mnist_fmnist}, showing the same trend as the causal signal strength increases.

Surprisingly, \emph{LightViT}'s $p$-value remained high despite the strength of the causal cue, suggesting a reliance on bucket-level structure over the simulated rotation. Prior Rotated MNIST studies---e.g., Ghifary et al.~\cite{ghifary2015domain} and Motiian et al.~\cite{motiian2017unified}---although in a different setting, also report unusually low performance specifically at $45^{\circ}$ due to increased class overlap. Consistent with this, we observe that particularly at $45^{\circ}$ rotation, \emph{LightViT} behaves unexpectedly with $p=0.3360$. These results indicate that \emph{LightViT} is not well-aligned with the rotational cue under study. More broadly, the observed $p$-values suggest that permutation-based significance testing may help identify whether a given model is suitable for a target causal cue. A deeper investigation of these effects is left for future work.


\subsubsection{Colored FashionMNIST}

We test whether a color cue enable binary discrimination on FashionMNIST. 
Classes \{T-shirt/Top, Trouser, Pullover, Dress, Coat\} are defined as \emph{treated} (\(T{=}1\)), 
and \{Sandal, Shirt, Sneaker, Bag, Ankle Boot\} as \emph{untreated} (\(T{=}0\)). 
Each grayscale image \(\mathbf{I}\!\in\![0,1]^{H\times W}\) is first replicated across RGB channels. 
Per-image channel multipliers are sampled from normal distributions:
\begin{align}
\text{If } T{=}1:\quad
&r \sim \mathcal{N}(\mu_R, \sigma_R^2), \\
&g, b \sim \mathcal{N}(\mu_F, \sigma_F^2); \\[4pt]
\text{If } T{=}0:\quad
&r, g, b \sim \mathcal{N}(\mu_F, \sigma_F^2) \text{ independently.}
\end{align}
The colorized image is generated by scaling the grayscale input:
\begin{equation}
(\mathbf{R}', \mathbf{G}', \mathbf{B}') = 
\mathrm{clip}\!\big(\mathbf{I} \odot (r, g, b),\, 0,\, 1\big),
\label{eq:colorcue}
\end{equation}
where \(\odot\) denotes elementwise multiplication. 
Here, \(\mu_F = 0.5\) and \(\sigma_F = 0.03\) are shared across the green and blue channels in both groups, while treatment strength is controlled by varying \(\mu_R\) over multiple  levels (\(\mu_R \in [0.55, 0.85],\ \Delta = 0.05\)). 
This process alters only the global red color intensity (which is the signal) per image, preserving spatial structure (refer Fig.\ref{fig:fashionmnist}). 

For each treatment level we perform all unique permutations, where each permutation retrains the models for 5 epochs to estimate the null accuracy distribution. As expected, when \(\mu_R = 0.5\) —no significant signal is present—the accuracy remains high on both models, but the \(p\)-value is also high, indicating no meaningful causal effect. 
As $\mu_R$ increases, accuracy exhibits an upward trend while the $p$-value decreases, suggesting stronger causal evidence (Table~\ref{tab:causal_mnist_fmnist}). Interestingly, \emph{LightViT} demonstrated same behavior as \emph{LightCNN} suggesting both architectures are trustworthy for the task.

\begin{table}[t]
\centering
\scriptsize
\caption{
Evaluation on controlled benchmarks using LightCNN and LightViT architectures. 
Observed accuracy is computed under true label assignments, and lower $p$-values indicate stronger causal evidence. 
Here, $p_{\min} = 0.0040$ for both experiments.
}
\label{tab:causal_mnist_fmnist}
\resizebox{\linewidth}{!}{
\begin{tabular}{llcccc}
\toprule
& & \multicolumn{2}{c}{\textbf{LightCNN}} & \multicolumn{2}{c}{\textbf{LightViT}} \\
\cmidrule(lr){3-4} \cmidrule(lr){5-6}
\textbf{Dataset} & \textbf{Treatment} & \textbf{Acc. (\%)} & \textbf{$p$-value} & \textbf{Acc. (\%)} & \textbf{$p$-value} \\
\midrule
\multirow{10}{*}{Rotated MNIST}
& $0^\circ$  & 98.51 & 0.9080 & 95.59 & 0.5360 \\
& $5^\circ$  & 98.88 & 0.7280 & 96.55 & 0.2960 \\
& $10^\circ$ & 99.70 & 0.0080 & 96.89 & 0.2360 \\
& $15^\circ$ & 99.73 & 0.0080 & 93.33 & 0.8560 \\
& $20^\circ$ & 99.71 & 0.0080 & 95.75 & 0.6040 \\
& $25^\circ$ & 99.73 & 0.0080 & 97.46 & 0.1400 \\
& $30^\circ$ & 99.82 & 0.0040 & 98.49 & 0.0040 \\
& $45^\circ$ & 99.88 & 0.0080 & 97.23 & 0.3360 \\
& $60^\circ$ & 99.96 & 0.0040 & 96.17 & 0.0880 \\
& $90^\circ$ & 99.86 & 0.0080 & 98.84 & 0.0120 \\
\midrule
\multirow{8}{*}{Colored FMNIST}
& $\mu_R=0.50$ & 93.66 & 0.4120 & 92.28 & 0.2440 \\
& $\mu_R=0.55$ & 94.92 & 0.1560 & 94.31 & 0.0840 \\
& $\mu_R=0.60$ & 97.52 & 0.0240 & 97.43 & 0.0120 \\
& $\mu_R=0.65$ & 99.16 & 0.0120 & 99.02 & 0.0080 \\
& $\mu_R=0.70$ & 99.80 & 0.0080 & 99.75 & 0.0120 \\
& $\mu_R=0.75$ & 99.96 & 0.0080 & 100.0 & 0.0040 \\
& $\mu_R=0.80$ & 100.0 & 0.0080 & 100.0 & 0.0080 \\
& $\mu_R=0.85$ & 100.0 & 0.0040 & 100.0 & 0.0080 \\
\bottomrule
\end{tabular}
}
\end{table}

\subsection{Real Biomedical Demonstrations}
\label{sec:biomedical}

\subsubsection{Bacteria Quantitative Phase Microscopy Dataset (QPM-WGS-AMR-21)}

We utilize the QPM-WGS-AMR-21~\cite{ahmad2023highly} dataset, which contains quantitative phase microscopy (QPM) images of 21 bacterial strains spanning five species: \textit{Acinetobacter baumannii}, \textit{Escherichia coli}, \textit{Klebsiella pneumoniae}, \textit{Staphylococcus aureus}, and \textit{Bacillus subtilis}. Each isolate is accompanied by whole genome sequencing (WGS) and antimicrobial resistance (AMR) profiles, enabling tasks such as Gram typing, AMR prediction, species classification, and strain discrimination. We evaluate three binary tasks constructed from this dataset. For AMR prediction (WT vs.\ NWT), we use ten isolates (five WT, five NWT) following the grouping in Ahmad et al.~\cite{ahmad2023highly}. The same ten isolates are used for Gram classification (GP vs.\ GN), comprising three Gram-positive and seven Gram-negative strains. Finally, we include a species-level task distinguishing \textit{Acinetobacter baumannii} (AB) from \textit{Klebsiella pneumoniae} (KP), using nine isolates (four AB, five KP) to provide a complementary pathogen-level benchmark.

For all tasks, we extract embeddings using a pretrained ResNet-18 backbone and train a lightweight classification head, following the setup in prior work~\cite{ahmad2023highly}. Class-balanced loss weighting is applied where appropriate. As shown in Table~\ref{tab:resnet18_results}, ResNet-18 achieves moderate --yet well above random chance--  accuracy on WT vs.\ NWT and GP vs.\ GN, but the high $p$-values indicate that these results are not trustworthy under our framework. This suggests that the model does not capture a strong or stable underlying signal for AMR or Gram-level discrimination. In contrast, the AB vs.\ KP task yields a lower $p$-value than the other two tasks, indicating a more reliable species-level signal; however, the value remains comparatively large relative to its own $p_{\min}$ (i.e. $1/M$ where $M$ is the number of permutations), meaning the evidence is still weak.

Overall, these results show that accuracy alone can be misleading. Significance testing using our framework provides an added layer of caution against spurious dataset-driven correlations.

\begin{table}[t]
\centering
\tiny
\setlength{\tabcolsep}{2pt}
\renewcommand{\arraystretch}{0.70}
\caption{
Evaluation of ResNet-18 on three binary classification tasks. 
Reported values show single-test accuracy and permutation-based $p$-values. 
Minimum attainable resolutions:
WT vs NWT ($p_{\min}=0.0040$), GP vs GN ($p_{\min}=0.0083$), AB vs KP ($p_{\min}=0.0080$).
}
\label{tab:resnet18_results}
\resizebox{0.50\linewidth}{!}{%
\begin{tabular}{lcc}
\toprule
Experiment & Acc. (\%) & $p$-value \\
\midrule
WT vs NWT & 72.00 & 0.3120 \\
GP vs GN  & 76.09 & 0.3403 \\
AB vs KP  & 77.04 & 0.0480 \\
\bottomrule
\end{tabular}
}
\end{table}

\subsubsection{Raman Bacteria-ID Dataset}

Raman spectroscopy is an emerging label-free technique increasingly used for biomedical applications, including pathogen detection. Here we use the public Raman bacteria-ID dataset from Ho et al.~\cite{ho2019rapid}, which contains Raman spectra of 30 bacterial and yeast isolates acquired with a 633\,nm laser on gold-coated silica substrates. Each isolate provides $\sim$2{,}000 low–signal-to-noise spectra, along with two clinical cohorts containing 400 and 100 spectra per isolate. Labels include species identity, empiric antibiotic group, and binary resistance status for methicillin-resistant \textit{Staphylococcus aureus} (MRSA) as methicillin-susceptible \textit{S. aureus} (MSSA), making this a large-scale benchmark for Raman-based bacterial classification~\cite{al2022highly,liu2023classification,liu2025deep,zhou2022ramannet}.

\begin{figure}[!h]
  \centering
  \includegraphics[width=\linewidth]{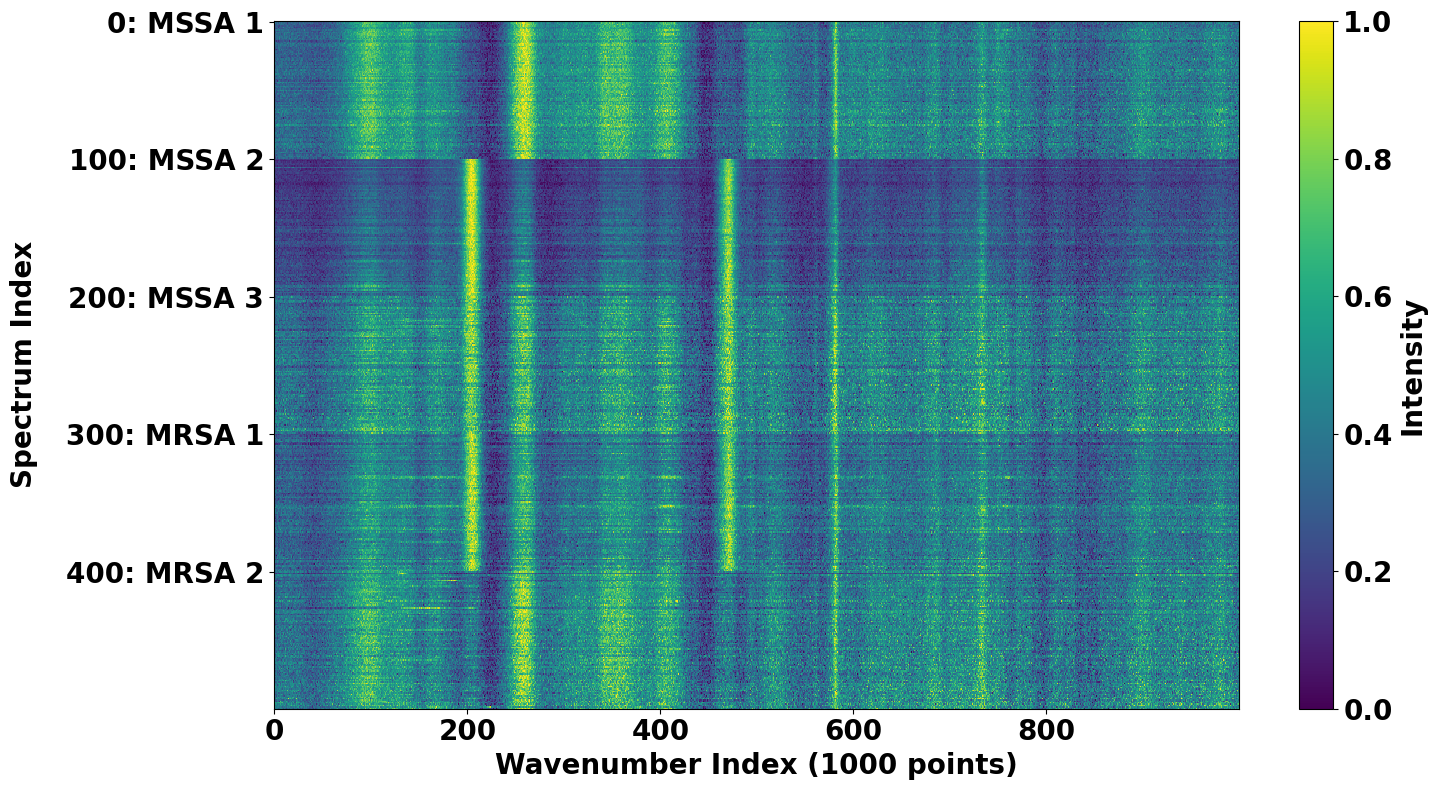}
  \caption{Representative Raman spectra for the MRSA vs.\ MSSA experiment. 
Each subclass (MRSA~1–2, MSSA~1–3) contains 100 spectra, with 1000 Raman shift points per spectrum. 
Spectral profiles between MRSA and MSSA appear visually similar, highlighting the challenge of discriminating antibiotic resistance from spectral data alone.}
  \label{fig:mrsa}
\end{figure}

We evaluate two binary tasks from this dataset: (i) Penicillin vs.\ Meropenem antibiotic grouping, and (ii) MRSA vs.\ MSSA resistance classification. The Penicillin vs.\ Meropenem task is challenging because each antibiotic group contains multiple isolates with highly similar Raman spectra, creating a high risk of overfitting to strain-level artifacts. We first trained a three-layer MLP baseline using 1000-dimensional Raman inputs. As shown in Table~\ref{tab:antibiotic_results}, this model achieved reasonable accuracy but yielded high $p$-values, indicating weak trustworthiness in our framework. We then evaluated an improved MLP with spectral derivatives, residual connections, LayerNorm, and GELU, along with the RamanNet architecture~\cite{zhou2022ramannet}(more model details in the supplementary). Both models produced higher accuracies and much lower $p$-values, demonstrating that the underlying antibiotic-group signal is detectable when a sufficiently expressive model is used. Thus, our framework distinguishes whether a negative result reflects non-IID structure or simply an underpowered model. For the MRSA vs.\ MSSA task, the Raman spectra for the two classes are nearly indistinguishable (Fig.~\ref{fig:mrsa}). Yet all models achieved well-above-chance accuracies (Table~\ref{tab:antibiotic_results}). But the corresponding $p$-values remained high across architectures (Table~\ref{tab:antibiotic_results}), indicating a consensus on the lack of a Raman signal indicative of AMR (related to MRSA/MSSA).

\begin{table}[t]
\centering
\footnotesize
\caption{Causal evaluation for Penicillin vs.\ Meropenem and MRSA vs.\ MSSA. 
Lower $p$-values indicate stronger causal significance. 
(Here, $p_{\min}=0.0002$ for Penicillin vs.\ Meropenem and $p_{\min}=0.1000$ for MRSA vs.\ MSSA.)}
\label{tab:antibiotic_results}
\resizebox{\linewidth}{!}{
\begin{tabular}{lcc|cc}
\toprule
\textbf{Model} &
\multicolumn{2}{c}{\textbf{Penicillin vs.\ Meropenem}} &
\multicolumn{2}{c}{\textbf{MRSA vs.\ MSSA}} \\
\cmidrule(lr){2-3} \cmidrule(lr){4-5}
& Acc. (\%) & $p$ & Acc. (\%) & $p$ \\
\midrule
Simple MLP        & 75.00 & 0.1050 & 80.20 & 0.8000 \\
Improved MLP      & 77.80 & 0.0010 & 73.80 & 1.0000 \\
RamanNet~\cite{zhou2022ramannet} & 80.14 & 0.0150 & 70.40 & 0.9000 \\
\bottomrule
\end{tabular}
}
\end{table}

Misclassifying  MRSA as MSSA in clinical practice can cause serious treatment failures, such as longer infections, higher mortality (20–50\% increase when therapy is delayed), and faster spread of resistance~\cite{cosgrove2003comparison,lodise2003outcomes}. On the other hand, labeling MSSA as MRSA can lead to unnecessary use of vancomycin, increasing the risk of kidney damage, \textit{C.~difficile} infections, and selection for vancomycin-resistant bacteria~\cite{dancer2008importance}. Because of these high clinical risks, trustworthiness frameworks like ours are critical to identify when and when not to trust future Raman-based diagnostics.

\section{Ablation Study}
\label{Ablation Study}


Training a model for each permutation is computationally expensive. We reduce this cost using ground-truth-free pretraining and by sampling a subset of permutations. We first assess how pretraining impacts convergence and runtime, and then evaluate how many permutations are needed for stable and efficient $p$-value estimation.


\subsection{Effect of Pretraining on Evaluation Efficiency}

As shown in Table~\ref{tab:pretrain_combined}, 
Masked Autoencoder (MAE)–based pretraining substantially reduces the total computation cost while maintaining comparable causal significance. 
In the Penicillin vs.\ Meropenem experiment, MAE pretraining achieved approximately a 40\% reduction in total evaluation time compared to training from scratch, with a consistent causal trend (\(p=0.009\)). 
A similar improvement is observed in the WT vs.\ NWT experiment, where self-supervised pretraining with ResNet-18 led to an estimated 95\% decrease in computation time. 
These results highlight that pretrained feature representations enable faster convergence and efficient permutation testing by eliminating the need for full model retraining, achieving over a 90\% overall reduction in evaluation time without compromising interpretability or causal reliability. Future work will further explore additional pretrained methods to enhance robustness and generalizability across diverse datasets.

\begin{table}[t]
\centering
\scriptsize
\caption{Comparison of pretrained and non-pretrained models for RamanNet (Penicillin vs.\ Meropenem) and ResNet-18 (WT vs.\ NWT). (Here, $p_{\min}$ = 0.0002 for Penicillin vs.\ Meropenem and $p_{\min}$ = 0.0083 for WT vs.\ NWT)}
\label{tab:pretrain_combined}

\resizebox{\columnwidth}{!}{%
\begin{tabular}{lcccc}
\toprule
\textbf{Experiment} & 
\textbf{Embed. time (min:s)} & 
\textbf{Perm. time (min:s)} & 
\textbf{Acc. (\%)} & 
\textbf{$p$-value} \\
\midrule

\textbf{Penicillin vs.\ Meropenem} \\
\cmidrule(lr){1-1}
Without Pretraining   & --      & $36:23$  & $80.14$ & $0.0160$ \\
With MAE Pretraining  & $0:38$  & $22:43$  & $73.67$ & $0.0100$ \\
\midrule

\textbf{WT vs.\ NWT} \\
\cmidrule(lr){1-1}
Without Pretraining   & --       & $651:55$ & $79.88$ & $0.6059$ \\
With MAE Pretraining  & $12:00$  & $11:27$  & $72.00$ & $0.3120$ \\
\bottomrule
\end{tabular}
}
\end{table}

\subsection{Number of Permutations for Stable Causal Inference}

The stability of causal inference depends on the number of random label permutations used to approximate the null distribution. 
Evaluating all permutations guarantees the exact Fisher-style $p$-value but is computationally expensive; therefore, we analyze how partial sampling behaves.

\begin{figure}[ht]
  \centering
  \includegraphics[width=\linewidth]{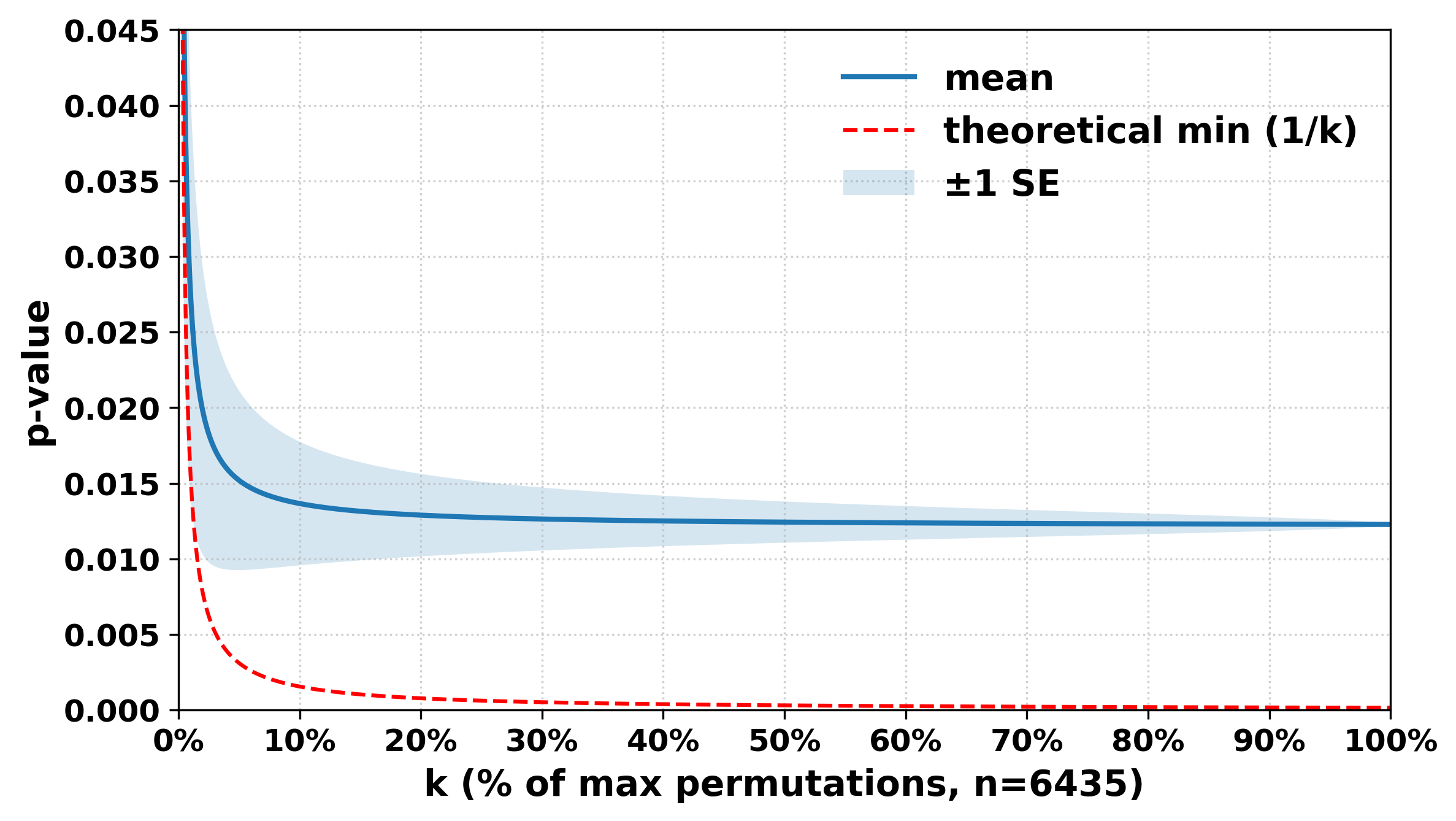}
  \caption{Convergence of causal $p$-value estimation with increasing number of sampled permutations.}
  \label{fig:perm_stability}
\end{figure}



For the Peniciline vs Meropenem task, we found that the estimated causal $p$-values rapidly converge even when using a small subset of permutations. 
Although the full null distribution contained 6{,}435 permutations, stable estimates were achieved with only 1{,}000 samples, producing $p$-values within the same significance range as full enumeration. 
As shown in Fig.~\ref{fig:perm_stability}, $p$-values stabilize after evaluating only $\sim 15.54\%$ of all permutations, enabling over $\sim 84.46\%$ time savings with no loss of statistical reliability.

\section{Conclusion}
\label{Conclusion}

We present a framework that redefines how trustworthiness in ML should be evaluated. Instead of relying solely on high test accuracy, our stochastic proof-by-contradiction approach provides a causal benchmark, requiring models to fail under data-hierarchy-aware randomized label permutations if they are truly learning meaningful patterns. By integrating data-hierarchy-aware permutations on the Rubin causal model, to calculate Fisher-style $p$-values, our method establishes a unified, interpretable standard that every ML study (especially bio-medical and clinical ones) can adopt to verify when IID assumptions are questionable. Our results across multiple datasets highlight clear guidelines for future evaluations that causal significance should accompany accuracy. Our framework thus serves as a reproducible, model-agnostic foundation for reliable and scientifically rigorous ML evaluation.

{
    \small
    \bibliographystyle{ieeenat_fullname}
    \bibliography{main}
}


\clearpage
\pagenumbering{arabic} 
\setcounter{page}{1}   
\setcounter{section}{0} 



\maketitlesupplementary

\thispagestyle{plain}

\section{Model Architectures Details}
\label{sup:model_details}

This section provides full architectural, training, and implementation details for the models used in the controlled benchmarks described in Sec.~\ref{sec:controlled}. 
All experiments were implemented in \texttt{PyTorch~2.0.1} and executed on NVIDIA RTX~6000 Ada Generation GPUs with CUDA~12.2 using mixed-precision training. 
Random seeds were fixed for all runs to ensure deterministic reproducibility.

\subsection{LightCNN Architecture}
\label{sup:lightcnn}

\textbf{Model Overview.}  
The \emph{LightCNN} is a compact convolutional neural network designed for low-resolution RGB inputs ($28\times28$) to provide a strong inductive baseline for local spatial learning. (used in Table~\ref{tab:causal_mnist_fmnist})

\textbf{Layer Configuration.}
\begin{itemize}[leftmargin=1.5em]
    \item \textbf{Input:} RGB image $(3\times28\times28)$.
    \item \textbf{Conv1:} $3 \rightarrow 32$ filters, kernel size $3\times3$, stride $1$, padding $1$; followed by ReLU and $2\times2$ MaxPool.
    \item \textbf{Conv2:} $32 \rightarrow 64$ filters, kernel size $3\times3$, stride $1$, padding $1$; followed by ReLU and $2\times2$ MaxPool.
    \item \textbf{Conv3:} $64 \rightarrow 128$ filters, kernel size $3\times3$, stride $1$, padding $1$; followed by ReLU and $2\times2$ MaxPool.
    \item \textbf{Flatten:} Output feature map $(128\times3\times3)$ flattened to a $1152$-D vector.
    \item \textbf{FC1:} Linear$(1152, 256)$, ReLU, Dropout($p=0.5$).
    \item \textbf{FC2:} Linear$(256, 2)$ for binary classification.
\end{itemize}

\textbf{Training Configuration.}
\begin{itemize}[leftmargin=1.5em]
    \item Optimizer: Adam
    \item Learning rate: $1\times10^{-4}$
    \item Batch size: 64
    \item Epochs: 5 per permutation
    \item Loss: CrossEntropyLoss
    \item Dropout: 0.5
\end{itemize}

\subsection{LightViT Architecture}
\label{sup:lightvit}

\textbf{Model Overview.}  
The \emph{LightViT} is a compact Vision Transformer adapted for $28\times28$ inputs. 
It employs a convolutional patch embedding stem to project local patches into a low-dimensional embedding space, followed by a lightweight Transformer encoder for global feature learning. (used in Table~\ref{tab:causal_mnist_fmnist})

\textbf{Patch Embedding.}
\begin{itemize}[leftmargin=1.5em]
    \item Patch size: $4\times4$, stride $4$
    \item Number of patches: $7\times7=49$
    \item Embedding dimension: $96$
    \item Operation: Conv2d$(3,96,\text{kernel}=4,\text{stride}=4)$ followed by flattening and transposing to $(B,49,96)$
\end{itemize}

\textbf{Positional and Classification Tokens.}
\begin{itemize}[leftmargin=1.5em]
    \item Learnable CLS token of shape $(1,1,96)$
    \item Learnable positional embedding $(1,50,96)$ (49 patches + 1 CLS)
    \item Dropout applied to combined embeddings ($p=0.1$)
\end{itemize}

\textbf{Transformer Encoder.}
\begin{itemize}[leftmargin=1.5em]
    \item Number of encoder blocks: 6
    \item Embedding dimension: 96
    \item Attention heads: 3
    \item MLP hidden dimension: $384$ ($96\times4$)
    \item Attention dropout: 0.1
    \item MLP dropout: 0.1
    \item Normalization: LayerNorm before attention and MLP
    \item Activation: GELU
\end{itemize}

\textbf{Classifier Head.}
\begin{itemize}[leftmargin=1.5em]
    \item CLS token representation → Linear$(96, 2)$ for binary prediction.
\end{itemize}

\textbf{Initialization.}
\begin{itemize}[leftmargin=1.5em]
    \item CLS and positional embeddings: truncated normal ($\sigma=0.02$)
    \item Linear and MLP layers: truncated normal ($\sigma=0.02$), zero bias
    \item Convolution layers: Kaiming Normal initialization
\end{itemize}

\textbf{Training Configuration.}
\begin{itemize}[leftmargin=1.5em]
    \item Optimizer: Adam
    \item Learning rate: $1\times10^{-4}$
    \item Batch size: 64
    \item Epochs: 5 per experiment or permutation
    \item Loss: CrossEntropyLoss
    \item Dropout: 0.1
\end{itemize}

\subsection{Simple MLP Architecture}

\textbf{Model Overview} - The \emph{Simple MLP} is a Multi-Layer Perceptron designed for binary classification tasks with high-dimensional input features. It employs a series of fully connected layers with batch normalization and dropout regularization to learn discriminative representations from input feature vectors. (used in Table~\ref{tab:antibiotic_results})

\textbf{Input Layer.}

\begin{itemize}[leftmargin=1.5em]
    \item Input dimension: $1000$
    \item Operation: Direct feature vector input
\end{itemize}

\textbf{Hidden Layers.}

\begin{itemize}[leftmargin=1.5em]
    \item Number of hidden layers: $3$
    \item Hidden dimensions: $[512, 256, 128]$
\item Each layer consists of:
\begin{itemize}
    \item Linear layer
    \item Batch normalization
    \item ReLU activation
    \item Dropout ($p = 0.1$)
\end{itemize}

    \item Layer sequence: $1000 \rightarrow 512 \rightarrow 256 \rightarrow 128$
\end{itemize}

\textbf{Output Layer.}

\begin{itemize}[leftmargin=1.5em]
    \item Final hidden representation $\rightarrow$ Linear$(128, 2)$ for binary prediction
\end{itemize}

\textbf{Initialization.}

\begin{itemize}[leftmargin=1.5em]
    \item Linear layers: Xavier uniform initialization
    \item Bias terms: Zero initialization
\end{itemize}

\textbf{Training Configuration.}

\begin{itemize}[leftmargin=1.5em]
    \item Optimizer: Adam
    \item Learning rate: $1\times10^{-4}$  (for MRSA vs MSSA), $5\times10^{-6}$ for (Penicillin vs.\ Meropenem)
    \item Batch size: $256$  (for MRSA vs MSSA), $64$ for (Penicillin vs.\ Meropenem)
    \item Epochs: $8$ per permutation
    \item Loss: CrossEntropyLoss
    \item Dropout: $0.1$
\end{itemize}

\subsection{Improved MLP Architecture}

\textbf{Model Overview} - The \emph{Improved MLP} is an enhanced Multi-Layer Perceptron designed for binary classification tasks with high-dimensional input features. It employs derivative feature extraction to capture first and second-order derivatives of the input signal, followed by residual feed-forward blocks with layer normalization for improved gradient flow and feature learning. (used in Table~\ref{tab:antibiotic_results})

\textbf{Feature Extraction.}

\begin{itemize}[leftmargin=1.5em]
    \item Input dimension: $1000$
    \item Derivative computation:
    \begin{itemize}
        \item First derivative: $\text{Conv1d}(x, [-1, 1])$ producing $dx$ of length $L-1$
        \item Second derivative: $\text{Conv1d}(x, [1, -2, 1])$ producing $d^2x$ of length $L-2$
        \item Original signal: center-cropped to length $L-2$ to match $d^2x$
    \end{itemize}
    \item Feature concatenation: $[x_{\text{cropped}}, dx_{\text{cropped}}, d^2x]$ flattened to $3 \times (L-2) = 2994$ dimensions
\end{itemize}

\textbf{Residual Feed-Forward Blocks.}

\begin{itemize}[leftmargin=1.5em]
    \item Number of blocks: $3$ (matching hidden dimensions)
    \item Hidden dimensions: $[512, 256, 128]$
\item Each \emph{ResidualFFN} block consists of:
\begin{itemize}
    \item Pre-normalization (LayerNorm)
    \item Linear layer
    \item GELU activation
    \item Dropout ($p = 0.2$)
    \item Second linear layer
    \item Residual connection (with projection when dimensions differ)
\end{itemize}

    \item Block sequence: $2994 \rightarrow 512 \rightarrow 256 \rightarrow 128$
\end{itemize}

\textbf{Classification Head.}

\begin{itemize}[leftmargin=1.5em]
    \item Layer normalization: LayerNorm$(128)$
    \item Final linear layer: Linear$(128, 2)$ for binary prediction
\end{itemize}

\textbf{Initialization.}

\begin{itemize}[leftmargin=1.5em]
    \item Linear layers: Xavier uniform initialization
    \item Bias terms: Zero initialization
    \item Projection layers (skip connections): Xavier uniform initialization
\end{itemize}

\textbf{Training Configuration.}

\begin{itemize}[leftmargin=1.5em]
    \item Optimizer: Adam
    \item Learning rate: $1\times10^{-4}$ (for MRSA vs MSSA), $1\times10^{-5}$ for (Penicillin vs.\ Meropenem)
    \item Batch size: $256$
    \item Epochs: $10$ per experiment or permutation
    \item Loss: CrossEntropyLoss
    \item Dropout: $0.2$ (for MRSA vs MSSA) , $0.5$ for (Penicillin vs.\ Meropenem)
    \item Weight decay: $1\times10^{-3}$
\end{itemize}

\section{Results Across Five Experimental Replications }

\begin{table}[!h]
\centering
\caption{
Evaluation on controlled benchmarks using LightCNN and LightViT architectures. 
Observed accuracy (Obs. Acc.) is computed under true label assignments, and lower $p$-values indicate stronger causal evidence against the null. 
Reported values are expressed as mean $\pm$ standard deviation across 5 randomized replications.( Here, $p_{\min}$ = 0.0040 for both experiments)
}
\label{tab:causal_mnist_fmnist22}
\resizebox{\linewidth}{!}{
\begin{tabular}{llccc}
\toprule
\textbf{Architecture} & \textbf{Dataset} & \textbf{Treatment Level} & \textbf{Obs. Acc. (\%)} & \textbf{$p$-value} \\
\midrule
\multirow{10}{*}{\textbf{LightCNN}} & Rotated MNIST & $0^\circ$  & $98.58 \pm 0.15$ & $0.8560 \pm 0.1080$ \\
&                      & $5^\circ$  & $99.21 \pm 0.57$ & $0.3600 \pm 0.4760$ \\
&                      & $10^\circ$ & $99.64 \pm 0.20$ & $0.0480 \pm 0.0920$ \\
&                      & $15^\circ$ & $99.71 \pm 0.05$ & $0.0080 \pm 0.0000$ \\
&                      & $20^\circ$ & $99.73 \pm 0.05$ & $0.0080 \pm 0.0040$ \\
&                      & $25^\circ$ & $99.76 \pm 0.02$ & $0.0040 \pm 0.0000$ \\
&                      & $30^\circ$ & $99.77 \pm 0.04$ & $0.0080 \pm 0.0000$ \\
&                      & $45^\circ$ & $99.88 \pm 0.03$ & $0.0040 \pm 0.0040$ \\
&                      & $60^\circ$ & $99.92 \pm 0.02$ & $0.0040 \pm 0.0040$ \\
&                      & $90^\circ$ & $99.92 \pm 0.02$ & $0.0040 \pm 0.0040$ \\
\midrule
\multirow{5}{*}{\textbf{LightCNN}}
& Colored FashionMNIST & $\mu_R=0.50$ & $93.67 \pm 0.01$ & $0.4080 \pm 0.0048$ \\
&                      & $\mu_R=0.55$ & $94.88 \pm 0.05$ & $0.1600 \pm 0.0065$ \\
&                      & $\mu_R=0.60$ & $97.42 \pm 0.16$ & $0.0240 \pm 0.0082$ \\
&                      & $\mu_R=0.65$ & $99.15 \pm 0.12$ & $0.0080 \pm 0.0040$ \\
&                      & $\mu_R=0.70$ & $99.82 \pm 0.02$ & $0.0040 \pm 0.0040$ \\
&                      & $\mu_R=0.75$ & $99.96 \pm 0.02$ & $0.0040 \pm 0.0040$ \\
&                      & $\mu_R=0.80$ & $100.00 \pm 0.00$ & $0.0040 \pm 0.0040$ \\
&                      & $\mu_R=0.85$ & $100.00 \pm 0.00$ & $0.0040 \pm 0.0040$ \\

\midrule
\multirow{10}{*}{\textbf{LightViT}} & Rotated MNIST & $0^\circ$  & $94.72 \pm 1.63$ & $0.6240 \pm 0.2160$ \\
&                      & $5^\circ$  & $94.17 \pm 2.80$ & $0.6480 \pm 0.2440$ \\
&                      & $10^\circ$ & $87.14 \pm 9.56$ & $0.7680 \pm 0.3240$ \\
&                      & $15^\circ$ & $91.07 \pm 4.02$ & $0.8800 \pm 0.0520$ \\
&                      & $20^\circ$ & $93.96 \pm 3.21$ & $0.6840 \pm 0.1960$ \\
&                      & $25^\circ$ & $94.59 \pm 2.80$ & $0.6280 \pm 0.3360$ \\
&                      & $30^\circ$ & $95.02 \pm 2.19$ & $0.6440 \pm 0.3680$ \\
&                      & $45^\circ$ & $95.14 \pm 1.88$ & $0.7000 \pm 0.2400$ \\
&                      & $60^\circ$ & $97.48 \pm 1.13$ & $0.2520 \pm 0.3160$ \\
&                      & $90^\circ$ & $99.13 \pm 0.22$ & $0.0080 \pm 0.0040$ \\
\midrule
\multirow{5}{*}{\textbf{LightViT}}
& Colored FashionMNIST & $\mu_R=0.50$ & $92.28 \pm 0.03$ & $0.2317 \pm 0.0072$ \\
&                      & $\mu_R=0.55$ & $94.37 \pm 0.13$ & $0.0800 \pm 0.0017$ \\
&                      & $\mu_R=0.60$ & $97.37 \pm 0.07$ & $0.0128 \pm 0.0016$ \\
&                      & $\mu_R=0.65$ & $99.10 \pm 0.05$ & $0.0064 \pm 0.0020$ \\
&                      & $\mu_R=0.70$ & $99.77 \pm 0.02$ & $0.0080 \pm 0.0000$
 \\
&                      & $\mu_R=0.75$ & $99.97 \pm 0.01$ & $0.0072 \pm 0.0040$ \\
&                      & $\mu_R=0.80$ & $100.00 \pm 0.00$ & $0.0080 \pm 0.0000$
 \\
&                      & $\mu_R=0.85$ & $100.00 \pm 0.00$ & $0.0080 \pm 0.0000$ \\
\bottomrule
\end{tabular}
}
\end{table}

\section{Ablation Study}

We observed that the p-values on Rotated MNIST (Table~\ref{tab:causal_mnist_fmnist}) were irregular and did not follow the expected progression across rotation angles. We think this may be because our original LightViT model was too strong for the task, producing saturated accuracies and very low permutation variance. To explore this hypothesis, we intentionally underpowered our LightViT backbone, resulting in the simplified UltraTinyViT, and repeated the evaluation and reported them in Table \ref{tab:causal_mnist_ultralightvit}.

\subsection{UltraTinyViT architecture.} Our UltraTinyViT is a heavily simplified ViT variant compared to the original LightVisionTransformer. First, we use a much smaller backbone: the embedding dimension is reduced from 96 to 16, the number of transformer blocks (depth) from 6 to 1, and the MLP expansion ratio from 4.0 to 2.0. The patch embedding is changed from a patch size of 4 (producing 49 patches arranged in a 7$\times$7 grid) to a patch size of 7 (producing 16 patches in a 4$\times$4 grid), and the convolutional projection is made bias-free. We completely remove the class token and learnable positional embeddings (and thus also positional interpolation and positional dropout) and instead perform simple mean pooling over all patch tokens before classification. The self-attention module is reduced from multi-head attention with three heads, a biased query-key-value projection, an explicit output projection, and dropout, to single-head attention with a bias-free query-key-value projection, no separate output projection, and no attention dropout. The MLP block is similarly simplified: both linear layers are bias-free, the expansion ratio is smaller, and all dropout is removed. Finally, all LayerNorm layers use non-affine normalization (no learnable scale or shift parameters), the classifier head does not include a bias term.

\begin{table}[t]
\centering
\scriptsize
\caption{
Evaluation on the Rotated MNIST benchmark using the UltraLightViT architecture.}
\label{tab:causal_mnist_ultralightvit}
\resizebox{\linewidth}{!}{
\begin{tabular}{llcc}
\toprule
& & \multicolumn{2}{c}{\textbf{UltraLightViT}} \\
\cmidrule(lr){3-4}
\textbf{Dataset} & \textbf{Treatment} & \textbf{Acc. (\%)} & \textbf{$p$-value} \\
\midrule
\multirow{10}{*}{Rotated MNIST}
& $0^\circ$  & 82.19 & 0.5880 \\
& $5^\circ$  & 83.71 & 0.4720 \\
& $10^\circ$ & 87.76 & 0.1760 \\
& $15^\circ$ & 88.85 & 0.1040 \\
& $20^\circ$ & 89.60 & 0.0520 \\
& $25^\circ$ & 90.65 & 0.0320 \\
& $30^\circ$ & 91.98 & 0.0160 \\
& $45^\circ$ & 93.22 & 0.0120 \\
& $60^\circ$ & 93.15 & 0.0160 \\
& $90^\circ$ & 93.36 & 0.0040 \\
\bottomrule
\end{tabular}
}
\end{table}

\subsection{Observation}

As seen in Table~\ref{tab:causal_mnist_ultralightvit}, the UltraLightViT results on the Rotated MNIST benchmark now exhibit the behavior we would intuitively expect from a well-calibrated causal test. At small rotation angles, the model already achieves high accuracies; however, the corresponding $p$-values also remain high. This combination indicates that, despite good predictive performance, the model has not yet learned a meaningful causal signal, as such accuracies are still compatible with the null distribution under label permutations. As the rotation angle increases, thereby strengthening the applied signal, the accuracies continue to rise while the $p$-values decrease in a smooth, nearly monotonic fashion, reaching their minimum at $90^\circ$. This stands in contrast to the earlier LightViT results, where the $p$-values fluctuated irregularly across treatments despite high accuracies. In our view, the UltraLightViT behavior is more interpretable: larger rotation angles make the observed accuracy stand further above the null distribution, resulting in smaller $p$-values and a clearer correspondence between treatment strength and causal significance.

Taken together, these results confirm our earlier hypothesis: the irregular $p$-value behavior in Table~1 was not due to a failure of the causal framework, but rather a consequence of using a model whose capacity was disproportionately high relative to the signal in the Rotated MNIST task. By reducing the model capacity, the permutation variance becomes more expressive, revealing the expected causal pattern. This finding highlights an important practical point of our framework: when a model is overly strong for a given dataset, accuracy alone can obscure the true causal structure, whereas the permutation-based $p$-values remain sensitive to such mismatches and provide a more reliable indicator of whether the model is responding to genuine treatment signals.






\end{document}